\pdfoutput=1 
\documentclass[runningheads]{llncs}
\usepackage{graphicx}
\usepackage{tabularx}
\usepackage{pdfpages}
\usepackage{multirow}
\usepackage{dsfont}
\usepackage{textcomp}

\usepackage{amsmath,amssymb} 
\usepackage{color}

\begin{document}

\title{Learning Monocular Depth by Distilling Cross-domain Stereo Networks}
\titlerunning{Learning Monocular Depth by Distilling Cross-domain Stereo Networks}
%
\author{Xiaoyang Guo\inst{1} \and
Hongsheng Li\inst{1}\thanks{Corresponding author} \and
Shuai Yi\inst{2} \and
Jimmy Ren\inst{2} \and
Xiaogang Wang\inst{1}
}
%
\authorrunning{X. Guo, H. Li, S. Yi, J. Ren and X. Wang}
%

\institute{CUHK-SenseTime Joint Laboratory, The Chinese University of Hong Kong \\
\email{\{xyguo,hsli,xgwang\}@ee.cuhk.edu.hk}
\and
SenseTime Research\\
\email{\{yishuai,rensijie\}@sensetime.com}
}
\maketitle              
%

\begin{abstract}
Monocular depth estimation aims at estimating a pixelwise depth map for a single image, which has wide applications in scene understanding and autonomous driving. Existing supervised and unsupervised methods face great challenges. Supervised methods require large amounts of depth measurement data, which are generally difficult to obtain, while unsupervised methods are usually limited in estimation accuracy. Synthetic data generated by graphics engines provide a possible solution for collecting large amounts of depth data. However, the large domain gaps between synthetic and realistic data make directly training with them challenging. 
In this paper, we propose to use the stereo matching network as a proxy to learn depth from synthetic data and use predicted stereo disparity maps for supervising the monocular depth estimation network. Cross-domain synthetic data could be fully utilized in this novel framework. Different strategies are proposed to ensure learned depth perception capability well transferred across different domains. Our extensive experiments show state-of-the-art results of monocular depth estimation on KITTI dataset.


\end{abstract}

\section{Introduction}
Depth estimation is an important computer vision task, which is a basis for understanding 3D geometry and could assist other vision tasks including object detection, tracking, and recognition. Depth can be recovered by varieties of methods, such as stereo matching~\cite{sgmstereo2008hirschmuller,dispnet2016mayer}, structure from motion~\cite{torr1999feature,agarwal2011building,wu2011visualsfm}, SLAM systems~\cite{orbslam2015mur,lsdslam2014engel,dtam2011newcombe}, and light field~\cite{tao2013depth}. Recently, monocular depth prediction from a single image~\cite{depth2014eigen,unsupervised2016garg,leftright2017godard} was investigated with deep Convolutional Neural Networks (CNN).

Deep CNNs could inherently combine local and global contexts of a single image to learn depth maps. The methods are mainly divided into two categories, supervised and unsupervised methods. For deep CNN based supervised methods~\cite{depth2014eigen,depthnormal2015eigen}, neural networks are directly trained with ground-truth depths, where conditional random fields (CRF) are optionally used to refine the final results. For unsupervised methods, the photometric loss is used to match pixels between images from different viewpoints by warping-based view synthesis. Some methods~\cite{unsupervised2016garg,leftright2017godard} learn to predict depth maps by matching stereo images, while some other ones~\cite{sfmlearner2017zhou,yang2017unsupervisedarxiv} learn depth and camera poses simultaneously from video frame sequences.

There are several challenges for existing monocular depth estimation methods. Supervised learning methods require large amounts of annotated data, and depth annotations need to be carefully aligned and calibrated. Ground truth captured by LIDAR is generally sparse, and structured light depth sensors do not work in strong light. Unsupervised learning methods~\cite{unsupervised2016garg,leftright2017godard} suffer from low texture, repeated pattern, and occlusions. It is hard to recover depth in occlusion regions with only the photometric loss because of the lack of cross-image correspondences at those regions.

Learning from synthetic data with accurate depth maps could be a potential way to tackle the above problems, but this requires synthetic data to be similar as realistic data in contents, appearance and viewpoints to ensure the model transferability. Otherwise, it is hard to adapt the model to realistic data due to the large domain gap. For example, a monocular depth estimation network pretrained with indoor synthetic data will have a bad performance in driving scenes, but it will perform better if pretrained with synthetic driving scene datasets like virtual KITTI~\cite{virtualkittiCVPR2016}. As a result, a lot of works are needed to build up corresponding synthetic datasets if the algorithm needs to be deployed in different scenes. On the other hand, we find that for state-of-the-art stereo matching algorithms~\cite{dispnet2016mayer,chang2018psmnet}, the stereo networks pretrained on cross-domain synthetic stereo images generalize much better to new domains compared with monocular depth networks, because the network learns the concept of matching pixels across stereo images instead of understanding high-level semantic meanings. Recently, stereo matching algorithms have achieved great success with the introduction of deep CNNs and synthetic datasets like Scene Flow datasets~\cite{dispnet2016mayer}, which inspires us to use stereo matching as a proxy task to learn depth maps from stereo image pairs, which can better utilize synthetic data and alleviate domain transfer problem compared with directly training monocular depth networks. 

In this paper, we propose a new pipeline for monocular depth learning with the guidance of stereo matching networks pretrained with cross-domain synthetic datasets. Our pipeline consists of three steps. First, we use a variant of DispNet~\cite{dispnet2016mayer} to predict disparity maps and occlusion masks with synthetic Scene Flow datasets. Then, the stereo matching network is finetuned with realistic data in a supervised or our novel unsupervised way. Finally, the monocular depth estimation network is trained under the supervision of the stereo network. 

Using stereo matching network as a proxy to learn depth has several advantages. On the one hand, stereo networks could efficiently make full use of cross-domain synthetic data and can be easier adapted to new domains compared to learning monocular depth. The synthetic datasets do not need to be separately designed for different scenes. On the other hand, the input data for stereo networks could be augmented by cropping and resizing to avoid over-fitting, while monocular depth networks usually fail to learn augmented images because it is sensitive to viewpoint changes. The experiment results show that our stereo matching network trained with synthetic data provides strong guidance for training monocular depth network, which could capture sharp boundaries and clear thin structures.

Our method achieves state-of-the-art results on the KITTI~\cite{kitti2013GeigerIJRR} dataset. Our contributions are as follows. 
1) We propose a novel monocular depth learning pipeline, which takes advantages of the power of stereo matching networks and synthetic data. By using stereo matching as a proxy task, the synthetic-to-realistic cross-domain problem could be effectively alleviated.
2) A novel unsupervised fine-tuning method is proposed based on the pretrained network to avoid occlusion problem and improve smoothness regularization. Visualization results show shaper boundaries and better occlusion predictions compared with previous unsupervised methods.
3) Our proposed pipeline achieves state-of-the-art monocular depth estimation performance in both unsupervised and semi-supervised settings.

\begin{figure}[t]
\centering
\includegraphics[width=\linewidth]{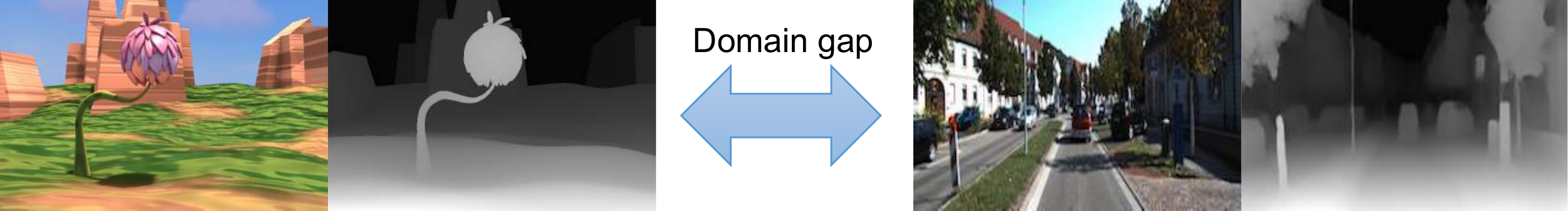}
\caption{Illustration of the large domain gap between synthetic and realistic data. Directly training monocular depth networks with cross-domain synthetic and realistic data results in inferior performance.}
\label{fig:domaingap}
\end{figure}

\section{Related Work}
Existing depth estimation methods can be mainly categorized into two types: estimation from stereo images and estimation from single monocular images. For depth from stereo images, disparity maps are usually estimated and then converted into depth maps with focal length and baseline (distance between two cameras).

{\bf Monocular Depth Estimation.\ }Monocular depth estimation aims to estimate depth values from a single image, instead of stereo images or multiple frames in a video. This problem is ill-posed because of the ambiguity of object sizes. However, humans could estimate the depth from a single image with prior knowledge of the scenes. Recently, learning based methods were explored to learn depth values by supervised or unsupervised learning. 

Saxena et al.~\cite{saxena2006learning,saxena20083} first incorporated multi-scale local and global features using Markov Random Field (MRF) to predict depth maps. Saxena et al.~\cite{saxena2009make3d} then extended to model locations and orientations of superpixels to obtain precise 3D structures from single images. Liu et al.~\cite{discretecontinuous2014liu} formulated depth estimation as a discrete-continuous graphic model.

Eigen et al.~\cite{depth2014eigen} first employed Convolutional Neural Networks (CNN) to predict depth in a coarse-to-fine manner and further improved its performance by multi-task learning~\cite{depthnormal2015eigen}. Liu et al.~\cite{liu2016learning} presented deep convolutional neural fields model by combining deep model with continuous CRF. Li et al.~\cite{depthnormalcrf2015li} refined deep CNN outputs with a hierarchical CRF. Multi-scale continuous CRF was formulated into a deep sequential network by Xu et al.~\cite{multi2017danxu} to refine depth estimation.

Unsupervised methods tried to train monocular depth estimation with stereo image pairs or image sequences and test on single images. Garg et al.~\cite{unsupervised2016garg} used novel image view synthesis loss to train a depth estimation network in an unsupervised way. Godard et al.~\cite{leftright2017godard} introduced left-right consistency regularization to improve the performance of view synthesis loss. Kuznietsov et al.~\cite{semidepth2017kuznietsov} combined supervised and unsupervised losses to further boost the performance. There were also works~\cite{demon2017ummenhofer,sfmlearner2017zhou,yang2017unsupervisedarxiv,sfmnetarxiv2017vijayanarasimhan,learningdirectarxiv2017wang} that try to recover depth maps and ego-motion from consecutive video frames.

{\bf Stereo Matching.\ }The target of stereo matching is to compute the disparity map given left-right image pairs. Usually, the input image pairs are rectified to make epipolar lines horizontal to simplify the matching problem. 

Stereo algorithms generally consist of all or some of the following four steps~\cite{scharstein2002taxonomy}: matching cost computation, cost aggregation, disparity optimization, and refinement. Local stereo methods~\cite{hosni2009local,kanade1994stereo,hosni2013fast} generally compute matching cost for all possible disparities and take the {\it winner-take-all} strategy to generate the final disparities. Global methods, including graph cut~\cite{kolmogorov2001graphcut}, belief propagation~\cite{sun2003stereobeliefpropagtion,yang2010constantspacebp} and semi-global matching (SGM)~\cite{sgmstereo2008hirschmuller}, take smoothness prior into consideration but are usually time-consuming. Higher level prior is investigated in Displets~\cite{displets2015guney}, which takes object shape into account in a superpixel-based CRF framework. 

Recently, deep learning has been successfully applied to stereo matching. Usually, these methods first train on large amounts of synthetic data, such as Scene Flow datasets~\cite{dispnet2016mayer}, and then finetune on realistic data. Zbontar and LeCun~\cite{mccnn2015zbontar} proposed to train a patch comparing network to compute the matching cost of stereo images. Mayer et al.~\cite{dispnet2016mayer} generalized the idea of Flownet~\cite{flownet2015dosovitskiy} and proposed a network structure with 1D correlation layer named DispnetC, which directly regresses disparity maps in an end-to-end way. Pang et al.~\cite{cascade2017pang} and Liang et al.~\cite{liang2018iresnet} extended DispnetC with cascade refinement structures. Recently, GC-Net~\cite{endtoend2017kendall} and PSMNet~\cite{chang2018psmnet} incorporated contextual information using 3-D convolutions over cost volume to improve the performance. 

For unsupervised learning, Zhou et al.~\cite{unsupervised2017zhou} presented a framework to learn stereo matching by iterative left-right consistency check. Tonioni et al.~\cite{unsupervised2017tonioni} combined off-the-shelf stereo algorithms and confidence measures to finetune stereo matching network.


\section{Method}
This section describes details of our method. For supervised monocular depth estimation methods~\cite{depth2014eigen,liu2016learning}, the accurate ground truth is usually limited and hard to obtain. For warping-based unsupervised methods~\cite{leftright2017godard,sfmlearner2017zhou}, the performance is usually limited due to the ambiguity of pixel matching. Directly training with synthetic data could only partly solve the problems, because it requires lots of works to design different synthetic
datasets for different scenes, and there is a large domain gap between synthetic and realistic data. Since the stereo matching network learns pixelwise matching instead of directly deducing depth from semantic features, it generalizes much better from synthetic domain to realistic domain compared with learning-based monocular depth methods. Inspired by the good generalization ability of stereo matching algorithms, we propose a novel pipeline for monocular depth learning to tackle the above limitations.

\begin{figure}[t]
\centering
\includegraphics[width=\linewidth]{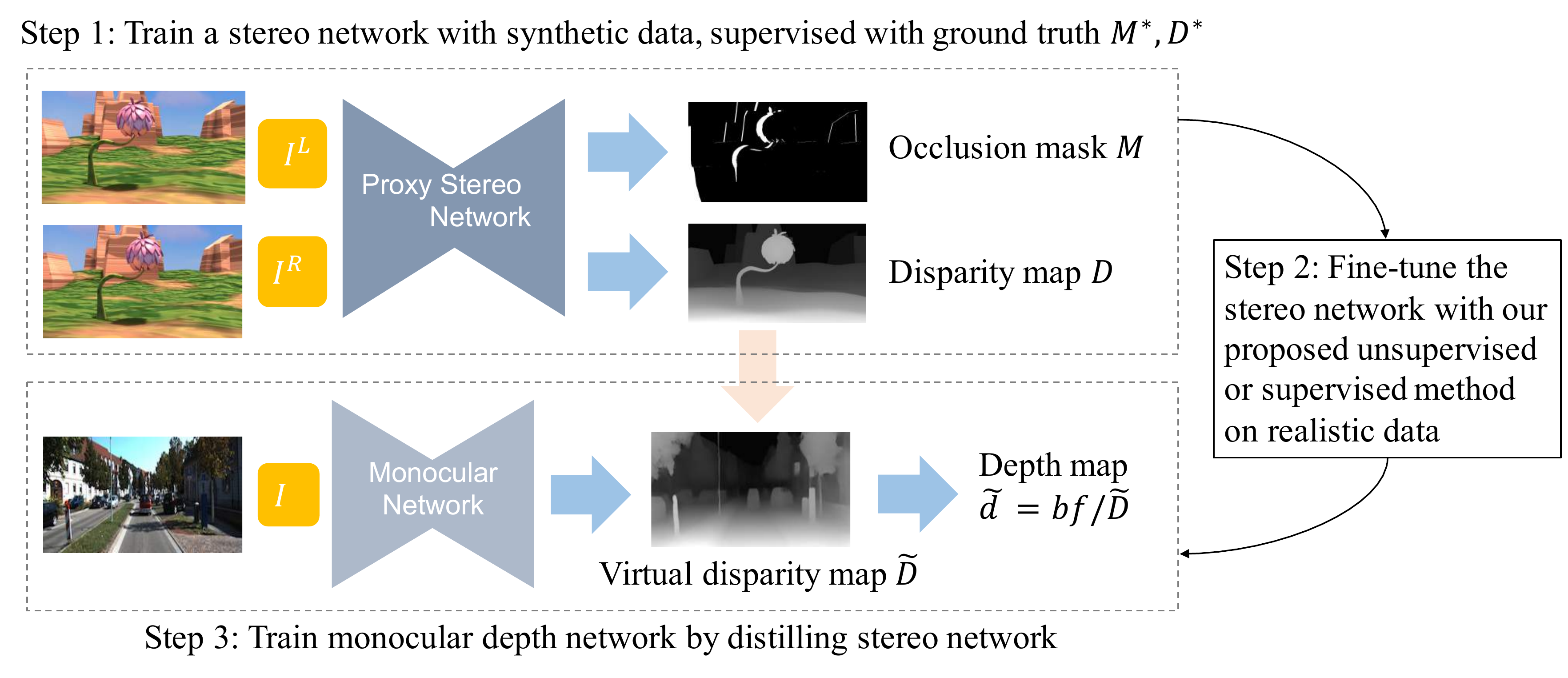}
\caption{The pipeline of our proposed method. The proxy stereo network is pre-trained with synthetic data and then finetuned on realistic data in supervised or unsupervised settings. The cross-domain gap problem could be mitigated by our stereo-to-monocular distillation approach.}
\label{fig:pipeline}
\end{figure}

Our method uses stereo matching as a proxy task for monocular depth learning and consists of three steps, as shown in Fig.~\ref{fig:pipeline}. In Sec.~\ref{sec:step1}, we train a stereo matching network with synthetic data to predict occlusion masks and disparity maps of stereo image pairs simultaneously. In Sec.~\ref{sec:step2}, the stereo network is finetuned on realistic data in supervised or unsupervised settings, depending on the availability of realistic data. In supervised settings, only 100 sampled training images are used to simulate the situation of limited ground truth. For unsupervised settings, a novel unsupervised loss is proposed to achieve better performance in occlusion regions. In Sec.~\ref{sec:step3}, we train the monocular depth estimation network by distilling the stereo network. In this way, with the proxy stereo network as a bridge from synthetic to realistic data, monocular depth estimation could better benefit from synthetic data, and much freedom is allowed for choosing synthetic data. 

\subsection{Training Proxy Stereo Matching Network with Synthetic Data}
\label{sec:step1}
Synthetic data can be used for pre-training to improve the performance of stereo matching. However, directly training monocular networks with synthetic data generally leads to inferior performance, because monocular depth estimation is sensitive to viewpoints and objects of the input scenes. Usually, synthetic datasets need to be carefully designed to narrow the domain gap, such as Virtual KITTI~\cite{virtualkittiCVPR2016} for driving scenes, which requires manually designing 3D scenes with large labor costs. However, we observe that stereo matching networks trained with only general-propose synthetic data can produce acceptable disparity map predictions on cross-domain realistic stereo image pairs. This inspires us to train stereo networks as a proxy to learn from synthetic data and a bridge between two data domains and two related tasks. 

\begin{figure}[!t]
    \centering
    \begin{tabular}{ccc}
    		\begin{tabular}{@{}c}\includegraphics[width=0.26\linewidth]{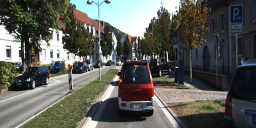}\end{tabular}
		  & 
		  \begin{tabular}{@{}c}\includegraphics[width=0.26\linewidth]{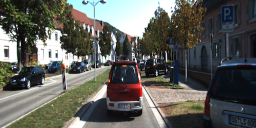}\end{tabular}
		  & 
		  \begin{tabular}{@{}c}\includegraphics[width=0.26\linewidth]{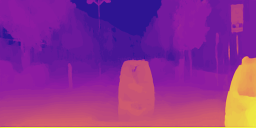}\end{tabular}
		  \\ 
		 (a) Left image & (b) Right image & (c) Disparity Map \\
		  		  \begin{tabular}{@{}c}\includegraphics[width=0.26\linewidth]{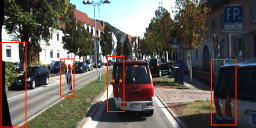}\end{tabular}
		  & 
		  		  \begin{tabular}{@{}c}\includegraphics[width=0.26\linewidth]{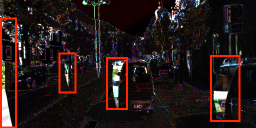}\end{tabular}
		  & 
		  \begin{tabular}{@{}c}\includegraphics[width=0.26\linewidth]{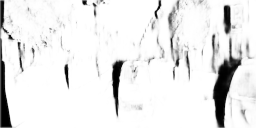}\end{tabular}
		  \\ 
		 (d) Warped right image & (e) Photometric error & (f) Occlusion mask
    \end{tabular}
\caption{Illustration of image warping and occlusion masks. The occlusion regions of the warped right image are not consistent with those of the left image even if the disparity map is correct. As a result, in the unsupervised fine-tuning, predicted occlusion masks are used to mask out the wrong supervision of photometric loss (see red rectangles) in occlusion regions.}
\label{fig:occlusion}
\end{figure}

We use a variant of DispnetC~\cite{dispnet2016mayer} as our proxy stereo matching network. DispnetC employs a 1-D correlation layer to extract the matching cost over all possible disparities and uses an encoder-decoder structure to obtain multiscale coarse-to-fine disparity predictions. Different from the original structure, our stereo network estimates multiscale occlusion masks in addition to disparity maps, as shown in Step 1 of Fig.~\ref{fig:pipeline}. Occlusion masks indicate whether the corresponding points of left image pixels are occluded in the right image. Fig.~\ref{fig:occlusion}(f) shows an example of occlusion masks. The occlusion masks will be used by our proposed unsupervised fine-tuning in Sec.~\ref{sec:step2} to avoid wrong photometric supervisions. 

The ground truth of occlusion masks is deduced from ground-truth disparity maps using left-right disparity consistency check,
\begin{align}
M^{*}_{ij}=
\mathds{1}\left(\left|D^{*L}_{ij}-D^{*wR}_{ij}\right|\leq1\right) =
\mathds{1}\left(\left|D^{*L}_{ij}-D^{*R}_{i(j-D^{*R}_{ij})}\right|\leq1\right)
,
\end{align} 
where the subscript $ij$ represents the value at the $i$th row and the $j$th column. $D^{*L/R}$ denotes the left/right disparity map, and $D^{*wR}$ is the right image which is warped to the left viewpoint. In occlusion regions, the values of $D^{*L}$ and $D^{*wR}$ are inconsistent. The threshold for the consistency check is set to 1-pixel length. The occlusion mask is set to 0 in occlusion regions and 1 in non-occlusion regions.

The loss is defined as $\mathcal{L}_\mathrm{stereo}=\sum_{m=0}^{M-1} w_m  \mathcal{L}_\mathrm{stereo}^m$ for M scales of predictions, where $w_m$ denotes the weighting factor for the $m$th scale. The loss function consists of two parts, disparity loss and occlusion mask loss,
\begin{align}
\mathcal{L}_\mathrm{stereo}^m=\mathcal{L}_{disp}+\mathcal{L}_{occ}
.
\end{align}
To train disparity maps, $L1$ regression loss is used to alleviate the impact of outliers and make the training process more robust. Occlusion masks are trained with binary cross-entropy loss as a classification task,
\begin{align}
\mathcal{L}_{occ}=
-\frac{1}{N}\sum_{i,j}M^*_{ij}\mathrm{log}({M}_{ij})+(1-M^*_{ij})\mathrm{log}{(1-{M}_{ij})}
,
\end{align}
where $N$ is the total number of pixels.

\subsection{Supervised and Unsupervised Fine-tuning Stereo Matching Network on Realistic Data}
\label{sec:step2}
The stereo matching network can be finetuned on realistic data depending on the availability of realistic depth data. In this paper, we investigate two ways to finetune the stereo matching network on realistic data, supervised learning with a very limited amount of depth data and our proposed unsupervised fine-tuning method.

{\bf Supervised Fine-tuning.} For supervised fine-tuning, only a multiscale $L1$ disparity regression loss is employed to refine the errors of pretrained models. The fine-tuning loss for the $m$th scale is defined as $\mathcal{L}_\text{stereo(supft)}^m=\mathcal{L}_{disp}$. Good results can be obtained with only a small amount of depth data, for example, 100 images. The stereo network can adapt from the synthetic domain to the realistic domain and fix most of the errors.

\begin{figure}[!t]
    \centering
    \begin{tabular}{cc}
    		\begin{tabular}{c}\includegraphics[width=0.3\linewidth]{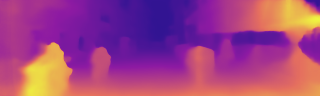}\end{tabular}
		  & 
		  \begin{tabular}{c}\includegraphics[width=0.3\linewidth]{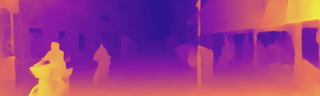}\end{tabular}
		  \\ 
		  (a) Finetuned with~\cite{leftright2017godard} & (b) Finetuned with proposed method \\
    \end{tabular}
\caption{Illustration of unsupervised finetuned predictions of stereo matching networks with the method of Godard et al.~\cite{leftright2017godard} and our proposed method.}
\label{fig:extrareg}
\end{figure}

{\bf Unsupervised Fine-tuning.} For unsupervised fine-tuning, we tried the unsupervised method of Godard et al.~\cite{leftright2017godard} to finetune our stereo networks with warping-based view synthesis loss but found that disparity predictions tended to become blurred, and the performance dropped, as shown in Fig.~\ref{fig:extrareg}(a). We argue that this is because of the matching ambiguity of the unsupervised loss and lack of occlusion handling. As a result, we propose to introduce additional occlusion handling terms and new regularization terms to improve the matching quality of unsupervised loss. 

Our proposed unsupervised loss requires the occlusion mask and disparity map predictions on realistic data from the un-finetuned stereo network, which is got in Sec.~\ref{sec:step1}. The predictions are denoted as ${M}_{un}$ and ${D}_{un}$ respectively. The unsupervised fine-tuning loss consists of three parts, the photometric loss, the absolute regularization term, and the relative regularization term,
\begin{align}
\mathcal{L}_\text{stereo(unsupft)}^m=\mathcal{L}_{photo}+\gamma_1\mathcal{L}_{abs}+\gamma_2\mathcal{L}_{rel}
,
\end{align}
where $\gamma_1$ and $\gamma_2$ are weighting factors. 

The photometric loss $\mathcal{L}_{photo}$ follows \cite{unsupervised2016garg,leftright2017godard} and uses warping-based view synthesis to learn disparity values by image reconstruction. The right image $I^R$ is first warped to the left viewpoint by bilinear sampling to obtain the warped right image $I^{wR}$, which is an estimate of the left image except for occlusion regions, which is illustrated in Fig.~\ref{fig:occlusion}(d). Since there is no pixelwise correspondence for pixels in occlusion regions, as shown in Fig.~\ref{fig:occlusion}(e), we use the occlusion mask ${M}_{un}$ to mask out the photometric supervision in occlusion regions to avoid incorrect supervisions. Then our photometric loss is given by
\begin{align}
\mathcal{L}_{photo}
=\frac{1}{N}\sum_{i,j}{M}_{un(ij)}\left|I^L_{ij}-I^{wR}_{ij}\right|
=\frac{1}{N}\sum_{i,j}{M}_{un(ij)}\left|I^L_{ij}-I^R_{i(j-{D}^L_{ij})}\right|
.
\end{align}

The absolute regularization term $\mathcal{L}_{abs}$ tries to make the newly-predicted disparity values close to the un-finetuned predictions especially in occlusion regions,
\begin{align}
\mathcal{L}_{abs}=\frac{1}{N}\sum_{i,j}\left(1-{M}_{un(ij)}+\gamma_3\right)\left|{D}^L_{ij}-{D}^{L}_{un(ij)}\right|
,
\end{align}
where ${D}_{un}$ is the disparity prediction of the un-finetuned stereo network, and $\gamma_3$ is a small regularization coefficient. 

The relative regularization term $\mathcal{L}_{rel}$ regularizes the prediction smoothness by the gradients of ${D}_{un}$, instead of input image gradients used in~\cite{leftright2017godard},
\begin{align}
\mathcal{L}_{rel}=
\frac{1}{N}\sum_{i,j}
\left|\nabla {D}^L_{ij}-\nabla {D}^{L}_{un(ij)}\right|
.
\end{align}

Fig.~\ref{fig:extrareg} compares the finetuned predictions using \cite{leftright2017godard} and our unsupervised fine-tuning method, which shows that our method is able to preserve sharp boundaries and accurate predictions in occlusion regions. In the experiment section, we will show that our method also achieves better quantitative results.

\subsection{Train Monocular Depth Network by Distilling Stereo Network}
\label{sec:step3}
By using a stereo network as a proxy, a large amount of synthetic data are fully utilized to train the stereo network without the cross-domain problem. To train the final monocular depth estimation network, we need to distill the knowledge of the stereo matching network. The monocular network also outputs multiscale predictions, and the loss is given by $\mathcal{L}_\mathrm{mono}=\sum_{m=0}^{M-1}w_m \mathcal{L}_{mono}^m$. For each scale, 
\begin{align}
\mathcal{L}_\mathrm{mono}^m
=\frac{1}{N}\sum_{i,j}\left|\tilde{D}^L_{ij}-{D}^{L}_{ij}\right|
,
\end{align}
where $\tilde{D}$ represents the virtual disparity maps predicted by monocular networks, and  ${D}$ denotes the predictions of the stereo matching network from Sec.~\ref{sec:step2}.

After getting the virtual disparity map $\tilde{D}$ from the monocular depth estimation network, the final depth map $\tilde{d}$ is then given by $\tilde{d}=bf/\tilde{D}$, where $b$ is the baseline distance between two cameras, and $f$ is the focal length of the lenses.

\section{Experiments}

\subsection{Datasets and Evaluation Metrics}
The synthetic Scene Flow Datasets~\cite{dispnet2016mayer} are used to pre-train our proxy stereo matching network, which is then finetuned on KITTI dataset~\cite{kitti2013GeigerIJRR} in supervised or unsupervised ways. The monocular network is then trained on the KITTI dataset under the supervision of the proxy stereo network.

{\bf Scene Flow datasets} are a collection of synthetic datasets, containing more than 39,000 stereo pairs for training and 4,000 for testing. The datasets are rendered by computer graphics in low cost and provide accurate disparity. The occlusion masks can be easily deduced from the ground-truth disparity maps by left-right consistency check. 

{\bf The KITTI dataset} is collected by moving vehicles in several outdoor scenes. We use the raw data of the KITTI dataset, which provide rectified stereo sequences, calibration information, and 3D LIDAR point clouds. Ground truth depth maps are inferred by mapping LIDAR points to the coordinate of the left camera. Our methods are all trained and tested on the Eigen split~\cite{depth2014eigen} of the KITTI dataset, which contains 22,600 image pairs for training, 888 for validation and 697 for testing. The Cityscapes dataset~\cite{Cordts2016Cityscapes} also provides driving scene stereo pairs and can be used to pre-train the monocular depth network.

{\bf Evaluation Metrics.} The evaluation metrics in \cite{depth2014eigen} are adopted to compare with the previous works. The metrics consist of absolute relative difference (Abs Rel), squared relative difference (Sq Rel), root mean square error (RMS), root mean square error in log scale (Log RMS), and threshold-based metrics. The threshold metric is defined as the percentage of pixels that satisfy $\delta={max}(d_{ij}/d_{ij}^*,d_{ij}^*/d_{ij})<thr$, where $d_{ij}$ and $d^*_{ij}$ are the depth prediction and depth ground truth for a pixel, and $thr$ is the threshold value. We cap the maximum depth values to 80 meters following previous works.

\subsection{Implementation Details}
\noindent{\bf Proxy stereo matching network.} We implement our proxy stereo matching network with the structure of DispNetC~\cite{dispnet2016mayer}, extra branch is added along with the disparity map output to predict occlusion masks. Equal weights are assigned for training disparity maps and occlusion masks. During training, the left-right image pairs are randomly flipped and then swapped to obtain the mirrored stereo pairs. Random resizing is then performed with a scaling factor between [0.8, 1.2] and followed by random cropping to generate stereo image patches. When performing image resizing, the corresponding ground-truth disparity values are multiplied by the scaling factor to ensure the correctness of stereo correspondences. Following~\cite{leftright2017godard}, the cropped images are then augmented by adjusting gamma in the range of [0.8, 1.2], illuminations in [0.8, 1.2], and color jitting in [0.95, 1.05], respectively. The network is optimized using Adam algorithm~\cite{kingma2014adam}. The parameters for Adam are set to $\beta_1$=0.9, $\beta_2$=0.999, and $\epsilon$=$10^{-8}$. The weighting factors $w_m$ for 4 multiscale predictions are set as $w_m=2^{-m}$ for both pre-training and fine-tuning, and $w_0$ corresponds to the final prediction.

When pre-training on Scene Flow datasets, the stereo network is trained for 50 epochs with a batch size of 4. The crop size for training is 768$\times$384, and original images without any augmentation are used for testing. The initial learning rate is $10^{-4}$ and is downgraded by half at epoch 20, 35, 45. The network is trained for two rounds to achieve better performance, following~\cite{cascade2017pang}. For each round, the learning rate restarts at $10^{-4}$. The stereo network trained with only synthetic data is denoted as {\it StereoNoFt}. 

Fine-tuning the stereo network is performed on the KITTI dataset. The input size for KITTI is 832$\times$256 for training and 1280$\times$384 for testing. Similar augmentation is performed on KITTI as those on Scene Flow datasets. The learning rate starts at $2$$\times$$10^{-5}$ and stops at $2.5$$\times$$10^{-6}$. For unsupervised fine-tuning, the whole training set of the Eigen Split is used, and the network is trained for 10 epochs. The weighting factors $\gamma_1$, $\gamma_2$ are set as 0.05 and 0.1, and $\gamma_3$ is 0.1. For supervised fine-tuning, only 100 images sampled from the training set are used to simulate the scenarios when ground truth depth maps are limited. The stereo models with unsupervised fine-tuning and supervised fine-tuning are denoted as {\it StereoUnsupFt} and {\it StereoSupFt100} respectively.

\begin{figure}[t]
\centering
\includegraphics[width=\linewidth]{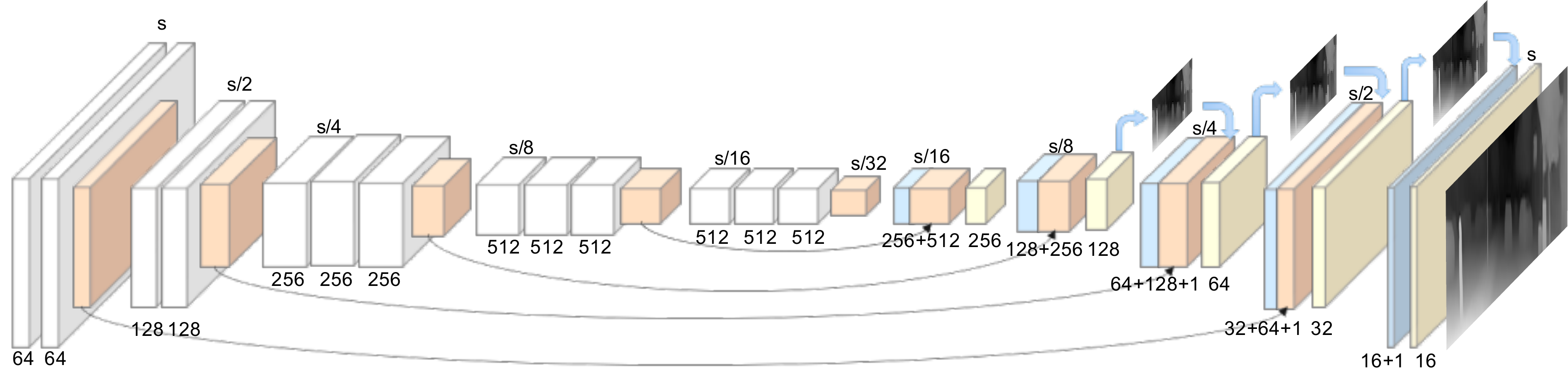}
\caption{The architecture of our monocular depth estimation network. The encoder is chosen as VGG-16 model. White blocks: 3$\times$3 convolution with ReLU. Red blocks: max pooling. Blue blocks: 4$\times$4 stride 2 deconvolution with Leaky ReLU. Yellow blocks: 3$\times$3 convolution with Leaky ReLU.}
\label{fig:mono-network}
\end{figure}

\noindent{\bf Monocular Depth Estimation Network.} The structure of our monocular depth estimation network follows \cite{leftright2017godard} with several important modifications. As shown in Fig.~\ref{fig:mono-network}, the encoder adopts the VGG-16~\cite{vgg16Simonyan14c} structure, and the decoder uses stacked deconvolutions and shortcut encoder features to recover multiscale depth predictions. The Leaky Rectified Linear Unit (LReLU)~\cite{lrelu-maas2013rectifier} with coefficient 0.1 is used in the decoder. The input images are randomly flipped and swapped. No cropping and scaling is used since the monocular depth estimation network is sensitive to viewpoint changes. The input images and disparity supervision from the stereo network are resized to 512$\times$256 to fit the input size of the monocular network. Gamma, illumination, and color augmentations are performed in the same way as the stereo network. The network is trained on the KITTI training set for 50 epochs with an initial learning rate of $10^{-4}$, which is then decreased by half at epoch 20, 35 and 45. If the monocular depth estimation model is supervised by {\it StereoNoFt}, the name for this model is denoted as {\it StereoNoFt\textrightarrow Mono}, and the naming rule is similar for other models. If the encoder of a model is initialized with weights pretrained for the ImageNet~\cite{ILSVRC15} classification task, the suffix {\it pt} is added to the model name.

\subsection{Comparison with State-of-the-arts}
\begin{figure}[!t]
    \centering 
    \begin{tabular}{@{}p{2.55cm}@{}c@{}c@{}c@{}c}
    	Input & 
		\begin{tabular}{@{}c}\includegraphics[width=0.26\linewidth]{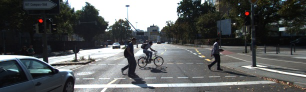}\end{tabular}
        &
        \begin{tabular}{@{}c}\includegraphics[width=0.26\linewidth]{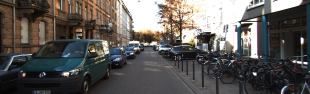}\end{tabular}
         &
        \begin{tabular}{@{}c}\includegraphics[width=0.26\linewidth]{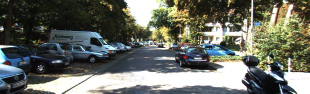}\end{tabular} 
        \\
		Ground Truth & 
		\begin{tabular}{@{}c}\includegraphics[width=0.26\linewidth]{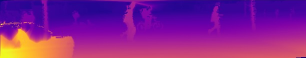}\end{tabular}
        &
        \begin{tabular}{@{}c}\includegraphics[width=0.26\linewidth]{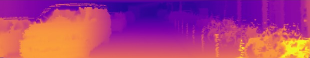}\end{tabular}
         &
        \begin{tabular}{@{}c}\includegraphics[width=0.26\linewidth]{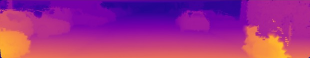}\end{tabular} 
        \\
		Eigen et al.~\cite{depth2014eigen} & 
		\begin{tabular}{@{}c}\includegraphics[width=0.26\linewidth]{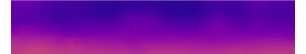}\end{tabular}
        &
        \begin{tabular}{@{}c}\includegraphics[width=0.26\linewidth]{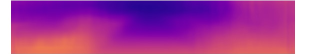}\end{tabular}
         &
        \begin{tabular}{@{}c}\includegraphics[width=0.26\linewidth]{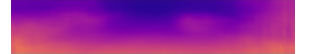}\end{tabular} 
        \\
        Godard et al.~\cite{leftright2017godard} & 
		\begin{tabular}{@{}c}\includegraphics[width=0.26\linewidth]{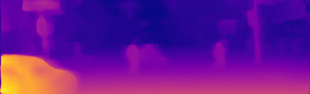}\end{tabular}
        &
        \begin{tabular}{@{}c}\includegraphics[width=0.26\linewidth]{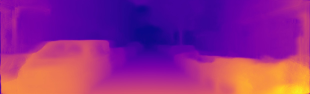}\end{tabular}
         &
        \begin{tabular}{@{}c}\includegraphics[width=0.26\linewidth]{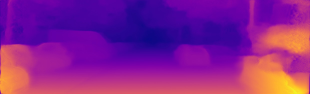}\end{tabular} 
        \\
		\begin{tabular}[l]{@{}l@{}}Kuznietsov \\et al.~\cite{semidepth2017kuznietsov}\end{tabular}
		 & 
		\begin{tabular}{@{}c}\includegraphics[width=0.26\linewidth]{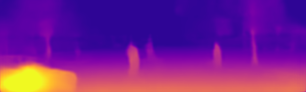}\end{tabular}
        &
        \begin{tabular}{@{}c}\includegraphics[width=0.26\linewidth]{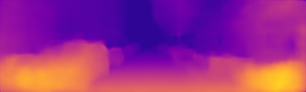}\end{tabular}
         &
        \begin{tabular}{@{}c}\includegraphics[width=0.26\linewidth]{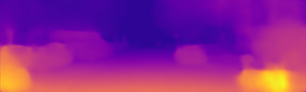}\end{tabular} 
        \\
         StereoNoFt \textrightarrow Mono & 
        \begin{tabular}{@{}c}\includegraphics[width=0.26\linewidth]{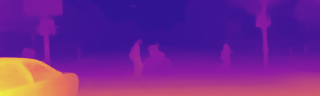}\end{tabular}
        &
        \begin{tabular}{@{}c}\includegraphics[width=0.26\linewidth]{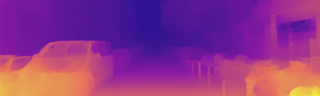}\end{tabular}
         &
        \begin{tabular}{@{}c}\includegraphics[width=0.26\linewidth]{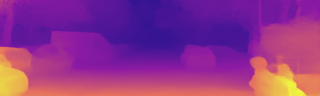}\end{tabular} 
        \\       	 
		 StereoUnsupFt \textrightarrow Mono & 
		        \begin{tabular}{@{}c}\includegraphics[width=0.26\linewidth]{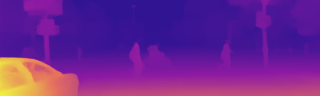}\end{tabular}
        &
        \begin{tabular}{@{}c}\includegraphics[width=0.26\linewidth]{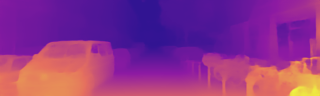}\end{tabular}
         &
        \begin{tabular}{@{}c}\includegraphics[width=0.26\linewidth]{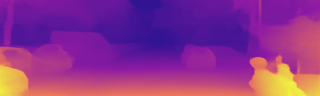}\end{tabular} 
        \\
		StereoSupFt100 \textrightarrow Mono & 
                \begin{tabular}{@{}c}\includegraphics[width=0.26\linewidth]{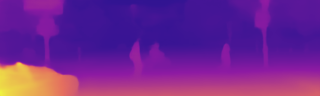}\end{tabular}
        &
        \begin{tabular}{@{}c}\includegraphics[width=0.26\linewidth]{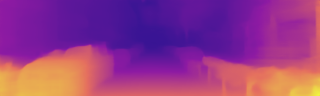}\end{tabular}
         &
        \begin{tabular}{@{}c}\includegraphics[width=0.26\linewidth]{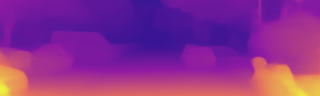}\end{tabular} 

    \end{tabular}
    \caption{Visualization of depth maps of different methods on KITTI test set. Results of~\cite{leftright2017godard} are shown without post-processing.}
    \label{fig:pred-visualization}
\end{figure}

In Table~\ref{table:results}, we compare the performance of our models with previous supervised and unsupervised methods. Depth results are visualized in Fig.~\ref{fig:pred-visualization} to compare with previous methods. We can see that our models can capture more detailed structures of the scenes. More visualization results can be found in the supplementary material.

First, we demonstrate the way we use stereo matching as a proxy task is crucial for utilizing cross-domain synthetic data. In the second part of Table~\ref{table:results}, we showed some results of monocular networks which were directly trained with synthetic data. We can get some conclusions. 1) The monocular network does not work on the KITTI dataset if only trained with Scene Flow datasets. Stereo networks generalize much better and have smaller synthetic-to-real domain transfer problems. 2) The performance of monocular networks has no apparent difference with or without pre-training on Scene Flow datasets. It is difficult for monocular depth networks to utilize cross-domain synthetic data by directly pre-training due to the large domain gap. 3) The monocular depth network requires a large amount of depth data to achieve good performance, while our method achieves better performance without or with only a limited number of depth maps. Therefore, our method, stereo-to-monocular distillation, could make better use of the large amount of cross-domain synthetic data to improve the performance of monocular depth estimation.

{\bf Comparison with Unsupervised Methods.} Our unsupervised model {\it StereoNoFt\textrightarrow Mono} and {\it StereoUnsupFt\textrightarrow Mono} are in similar settings with previous unsupervised methods~\cite{leftright2017godard} since they do not use ground truth depth of realistic data, and synthetic data can be obtained free of charge by rendering engines. From Table~\ref{table:results}, we can see that the monocular depth model supervised by the un-finetuned proxy stereo network ({\it StereoNoFt\textrightarrow Mono}) even surpasses the state-of-the-art unsupervised method~\cite{leftright2017godard}, which is based on warping-based view synthesis and difficult to obtain sharp boundaries and handle occlusion parts. In contrast, our method can take advantages of synthetic data to provide rich depth supervisions and infer the depth values in occlusion parts.  We also observe that monocular depth estimation can benefit from ImageNet pretrained weights (with postfix {\it pt} in experiment names) and pre-training on street scene datasets like the Cityscapes dataset under the supervision of the stereo model {\it StereoNoFt}. Our best unsupervised model {\it StereoUnsupFt\textrightarrow Mono pt} trained on Cityscapes and KITTI outperforms the method of~\cite{leftright2017godard} by 0.62 meters in terms of root mean squared error and decreases the squared relative error by 21.7\%.

{\bf Comparison with Supervised Methods. } For supervised fine-tuning, instead of using all ground truth in the training set of KITTI, our model {\it StereoSupFt100\textrightarrow Mono} only used 100 sampled images with LIDAR depths in the training set for fine-tuning. Results show that our method also works when only a limited number of accurate depth maps is available. We compare our results with the method of Kuznietsov et al.~\cite{semidepth2017kuznietsov}, which combines supervised and unsupervised losses to estimate monocular depth. However, although their method uses all ground truth data, we can see from Table~\ref{table:results} that all evaluation metrics of our model {\it StereoSupFt100\textrightarrow Mono pt},  surpass~\cite{semidepth2017kuznietsov} and all other previous supervised methods. We also provide the results of fine-tuning the proxy stereo network with all supervised data ({\it StereoSupFtAll\textrightarrow Mono}) for reference.

We also evaluate with a maximum depth cap of 50 meters in the bottom part of Table~\ref{table:results}. Similarly, our method beats all of the state-of-the-art methods.

\setlength{\tabcolsep}{0.8mm}
\begin{table}[t]  
\begin{center}
\caption{Quantitative results on KITTI~\cite{kitti2013GeigerIJRR} using the split of Eigen et al.~\cite{depth2014eigen}. All results are evaluated with the crop of~\cite{unsupervised2016garg} except \cite{depth2014eigen}. The results of \cite{depth2014eigen,liu2016learning} are from the paper of \cite{leftright2017godard} for fair comparision. $Sup.$ means supervised and 100 represents only using 100 ground truth depth maps. S, C, K denote Scene Flow datasets~\cite{dispnet2016mayer}, Cityscapes dataset~\cite{Cordts2016Cityscapes} and KITTI dataset~\cite{kitti2013GeigerIJRR} respectively. $res$ denotes using Resnet~\cite{he2016deep} as the encoder structure, and $pt$ denotes using model weights pretrained on ImageNet~\cite{ILSVRC15}.}
\label{table:results}
\scriptsize
\begin{tabular}{|l|c|c||c|c|c|c|c|c|c|} 
\hline
\multirow{3}{*}{Method} & \multirow{3}{*}{Sup.} & \multirow{3}{*}{Dataset} & \multicolumn{4}{c|}{lower is better} & \multicolumn{3}{c|}{higher is better} \\ \cline{4-10} 
                        &                       &                          & Abs     & Sq      & RMS     & Log    & $\delta<$   & $\delta<$  & $\delta<$  \\ 
                        &                       &                          & Rel     & Rel     &         & RMS    & $1.25$      & $1.25^2$   & $1.25^3$   \\ 
\hline
Eigen et al. Fine~\cite{depth2014eigen}& Yes & K & 
0.203 & 1.548 & 6.307 & 0.282 & 0.702 & 0.890 & 0.958\\
DCNF-FCSP FT~\cite{liu2016learning}  & Yes & K & 
0.201 & 1.584 & 6.471 & 0.273 & 0.680 & 0.898 & 0.967 \\
Godard et al.~\cite{leftright2017godard} & No & K & 
0.148 & 1.344 & 5.927 & 0.247 & 0.803 & 0.922 & 0.964 \\
Godard et al.~\cite{leftright2017godard} res pp & No & C,K & 
0.114 & 0.898 & 4.935 & 0.206 & 0.861 & 0.949 & 0.976 \\
Zhou et al.~\cite{sfmlearner2017zhou} & No & K & 
0.208 & 1.768 & 6.856 & 0.283 & 0.678 & 0.885 & 0.957 \\
Kuznietsov et al.~\cite{semidepth2017kuznietsov} res pt & Yes & K & 
0.113 & 0.741 & 4.621 & 0.189 & 0.862 & 0.960 & 0.986 \\
\hline
Direct Sup. & No & S & 0.662 & 9.502 & 16.03 & 1.407 & 0.053 & 0.124 & 0.209 \\
Direct Sup. (All of K) & Yes & K & 
0.105 & 0.717 & 4.422 & 0.183 & 0.874 & 0.959 & 0.983\\
Direct Sup. pt(S) (All of K) & Yes & S,K &
0.106 & 0.723 & 4.506 & 0.185 & 0.871 & 0.958 & 0.983 \\
Direct Sup. (100 of K) & 100 & K &
0.187 & 1.563 & 6.283 & 0.273 & 0.732 & 0.889 & 0.953 \\
Direct Sup. pt(S) (100 of K) & 100 & S,K &
0.194 & 1.560 & 6.001 & 0.267 & 0.737 & 0.896 & 0.956 \\
\hline
StereoNoFt\textrightarrow Mono & No & S\textrightarrow K & 
0.109 & 0.822 & 4.656 & 0.192 & 0.868 & 0.958 & 0.981 \\
StereoUnsupFt\textrightarrow Mono & No & S,K\textrightarrow K & 
0.105 & 0.811 & 4.634 & 0.189 & 0.874 & 0.959 & 0.982\\
StereoUnsupFt\textrightarrow Mono pt & No & S,K\textrightarrow K & 
0.099 & 0.745 & 4.424 & 0.182 & 0.884 & 0.963 & 0.983\\
StereoUnsupFt\textrightarrow Mono pt & No & S,K\textrightarrow C,K & 
\bfseries 0.095 & \bfseries 0.703 & \bfseries 4.316 & \bfseries 0.177 & \bfseries 0.892 & \bfseries 0.966 & \bfseries 0.984\\
\hline
StereoSupFt100\textrightarrow Mono pt & 100 & S,K\textrightarrow K & 
0.101  & 0.690 & 4.254 & 0.173 & 0.884 & 0.966 & \bfseries 0.986\\
StereoSupFtAll\textrightarrow Mono pt & Yes & S,K\textrightarrow K & 
0.097 & 0.653 & 4.170 & 0.170 & 0.889 & \bfseries 0.967 & \bfseries 0.986\\
StereoSupFt100\textrightarrow Mono pt & 100 & S,K\textrightarrow C,K & 
\bfseries 0.096  & \bfseries 0.641  & \bfseries 4.095  & \bfseries 0.168  & \bfseries 0.892  & \bfseries 0.967  & \bfseries \bfseries 0.986\\

\hline
\multicolumn{10}{c}{cap 50m} \\
\hline
Garg et al. [16] L12 Aug 8x & No & K &
0.169 & 1.080 & 5.104 & 0.273 & 0.740 & 0.904 & 0.962 \\
Godard et al.~\cite{leftright2017godard} res pp & No & C,K & 
0.108 & 0.657 & 3.729 & 0.194 & 0.873 & 0.954 & 0.979 \\
Kuznietsov et al.~\cite{semidepth2017kuznietsov} res pt & Yes & K & 
 0.108 & 0.595 & 3.518 & 0.179 &  0.875 & 0.964  & 0.988 \\
 \hline
StereoUnsupFt\textrightarrow Mono pt & No & S,K\textrightarrow K & 
0.094 & 0.555 & 3.347 & 0.172 & 0.895 & 0.968 & 0.985\\
StereoUnsupFt\textrightarrow Mono pt & No & S,K\textrightarrow C,K & 
\bfseries 0.090 & \bfseries 0.522 & \bfseries 3.258 & \bfseries 0.168 & \bfseries 0.902 & \bfseries 0.969 & \bfseries 0.986\\
\hline
StereoSupFt100\textrightarrow Mono pt & 100 & S,K\textrightarrow K & 
0.097 & 0.549 & 3.259 & 0.164 & 0.895 & 0.970 & \bfseries 0.988\\
StereoSupFt100\textrightarrow Mono pt & 100 & S,K\textrightarrow C,K & 
\bfseries 0.092 & \bfseries 0.515 & \bfseries 3.163 & \bfseries 0.159 & \bfseries 0.901 & \bfseries 0.971 & \bfseries 0.988\\
\hline

\end{tabular}
\end{center}
\end{table}

\subsection{Analysis of Fine-tuning for Proxy Stereo Models}
We first compared our proposed unsupervised fine-tuning method with the unsupervised method of \cite{leftright2017godard} ({\it StereoUnsupFt} and {\it StereoUnsupFt(\cite{leftright2017godard})} in Table~\ref{table:stereo-val-results}). The method of \cite{leftright2017godard} didn't explicitly handle the occlusion regions, so the predictions in the occlusion parts tended to be blurred. Our unsupervised fine-tuning method removed the incorrect supervision in occlusion regions and improved the regularization terms. From Fig.~\ref{fig:extrareg} and Table~\ref{table:stereo-val-results}, we can see our unsupervised fine-tuning method surpasses the method of \cite{leftright2017godard} quantitatively and qualitatively.

We then compare the performance of our stereo models {\it StereoNoFt}, {\it StereoUnsupFt}, {\it StereoSupFt100}, and corresponding distilled monocular depth estimation models. By comparing {\it StereoUnsupFt} and {\it StereoSupFt100} with {\it StereoNoFt}, we can see that both supervised and unsupervised fine-tuning improve the performance on the KITTI dataset. Although only 100 images are used for supervised fine-tuning, it still surpasses the model with unsupervised fine-tuning. The
performance of monocular depth networks improves with the performance of corresponding proxy stereo networks. Fine-tuning of stereo networks brings performance improvement for monocular depth estimation. The performance of our pipeline improves if the proxy stereo network is replaced with PSMNet~\cite{cascade2017pang}, and detailed results are reported in the supplementary material. As a result, if more advanced stereo matching networks and better fine-tuning strategies are employed, the performance of our method could be further improved.

\setlength{\tabcolsep}{0.7mm}
\begin{table}
\begin{center}
\caption{Comparison of our proxy stereo models and corresponding monocular depth models on the KITTI dataset (Eigen split) and Scene Flow datasets.}
\label{table:stereo-val-results}
\scriptsize
\begin{tabular}{|l|c|c|c|c|c|c|c|c|c|c|} 
\hline
\multirow{3}{*}{Method} & \multicolumn{7}{c|}{KITTI (Eigen split)}     & \multicolumn{3}{c|}{Scene Flow} \\ 
\cline{2-11}
                        & Abs & Sq  & RMS & Log & $\delta<$ & $\delta<$ & $\delta<$ & MAE  & \textgreater1px & \textgreater3px \\ 
                        & Rel & Rel &     & RMS & $1.25$    & $1.25^2$  & $1.25^3$  & (px) &     \%    &  \%      \\ \cline{1-8}
\hline
StereoNoFt & 
0.072 & 0.665 & 3.836 & 0.153 & 0.936 & 0.973 & 0.986 & 
\bfseries 3.41 & \bfseries 0.241 & \bfseries 0.096  \\
StereoUnsupFt(\cite{leftright2017godard}) & 	
0.076 & 0.691 & 4.076 & 0.173 & 0.915 & 0.960 & 0.979 & 
13.3 & 0.515 & 0.365 \\
StereoUnsupFt & 	
\bfseries 0.061 & 0.612 & 3.553 & 0.144 & \bfseries 0.948 & 0.975 & 0.986  & 
4.23 & 0.293 & 0.124 \\
StereoSupFt100 &
0.063 & \bfseries 0.562 & \bfseries 3.325 & \bfseries 0.137 & \bfseries 0.948 & \bfseries 0.978 & \bfseries 0.988 &
5.27 & 0.430 & 0.181 \\
\hline
StereoNoFt\textrightarrow Mono & 
0.109 & 0.822 & 4.656 & 0.192 & 0.868 & 0.958 & 0.981 &
- & - & -\\
StereoUnsupFt(\cite{leftright2017godard})\textrightarrow Mono & 
0.116 & 0.877 & 4.996 & 0.210 & 0.845 & 0.943 & 0.975 & 
- & - & -\\
StereoUnsupFt\textrightarrow Mono &
\bfseries 0.105 & 0.811 & 4.634 & 0.189 & \bfseries 0.874 & \bfseries 0.959 & 0.982 &
- & - & -\\
StereoSupFt100\textrightarrow Mono & 
0.111 & \bfseries 0.771 & \bfseries 4.449 & \bfseries 0.185 & 0.868 & 0.958 & \bfseries 0.983 & 
- & - & - \\
\hline
\end{tabular}
\end{center}
\end{table}

%

\subsection{Generalization to Other Datasets}
\begin{figure}[!t]
    \centering
    \footnotesize
    
    \begin{tabular}{cc@{}c@{}c@{}c}
		Make3d & 
        \begin{tabular}{@{}c}
            \includegraphics[width=0.2\linewidth]{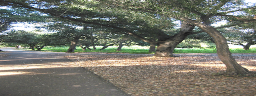}
        \end{tabular}
        &
        \begin{tabular}{@{}c}
            \includegraphics[width=0.2\linewidth]{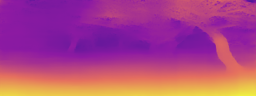}
        \end{tabular}
         &
        \begin{tabular}{@{}c}
            \includegraphics[width=0.2\linewidth]{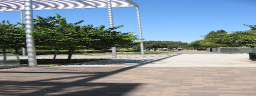}
        \end{tabular}
         &
        \begin{tabular}{@{}c}
            \includegraphics[width=0.2\linewidth]{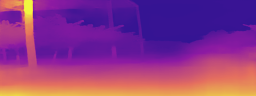}
        \end{tabular}\\ 
        Cityscapes & 
        \begin{tabular}{@{}c}
            \includegraphics[width=0.2\linewidth]{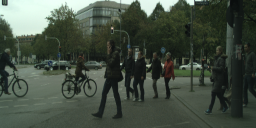}
        \end{tabular}
        &
        \begin{tabular}{@{}c}
            \includegraphics[width=0.2\linewidth]{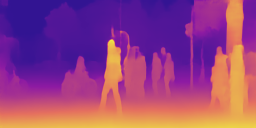}
        \end{tabular}&
        \begin{tabular}{@{}c}
            \includegraphics[width=0.2\linewidth]{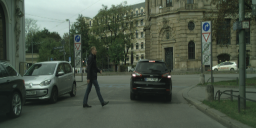}
        \end{tabular}
        &
        \begin{tabular}{@{}c}
            \includegraphics[width=0.2\linewidth]{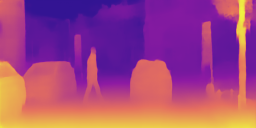}
        \end{tabular}
    \end{tabular}
    \caption{Qualitative depth estimation results on Make3D dataset~\cite{saxena2009make3d} and Cityscapes dataset~\cite{Cordts2016Cityscapes} predicted by the model {\it StereoUnsupFt\textrightarrow Mono}.}
    \label{fig:otherdatasets}
\end{figure}

We also tested our model {\it StereoUnsupFt\textrightarrow Mono} (Finetuned on the KITTI dataset) on Make3D dataset and Cityscapes dataset to show the generalization ability of the monocular depth model to other datasets. Some visualization results are shown in Fig.~\ref{fig:otherdatasets}.

\section{Conclusion}
In this paper, we have proposed a new pipeline for monocular depth estimation to tackle the problems existing in previous supervised and unsupervised methods. Our method utilizes deep stereo matching network as a proxy to learn depth from synthetic data and provide dense supervision for training monocular depth estimation network. Multiple strategies for fine-tuning the proxy stereo matching network are investigated, and the proposed unsupervised fine-tuning method successfully keeps detailed structures of predictions and improves the final performance. Our extensive experimental results show state-of-the-art results of monocular depth estimation on the KITTI dataset.

For future works, more advanced stereo matching algorithms and better fine-tuning strategies can be investigated to further improve the performance of the pipeline. Confidence measurement is another way to filter the noises of the outputs of stereo matching networks and provide better supervision for training monocular depth network.

\section*{Acknowledgements}
This work is supported by SenseTime Group Limited, the General Research Fund sponsored by the Research Grants Council of Hong Kong (Nos. CUHK14213616, CUHK14206114, CUHK14205615, CUHK14203015, CUHK14239816, CUHK419412, CUHK14207814, CUHK14208417, CUHK14202217), the Hong Kong Innovation and Technology Support Program (No.ITS/121/15FX).


\bibliographystyle{splncs04}
\bibliography{paper}

\begin{thebibliography}{10}
\providecommand{\url}[1]{\texttt{#1}}
\providecommand{\urlprefix}{URL }
\providecommand{\doi}[1]{https://doi.org/#1}

\bibitem{agarwal2011building}
Agarwal, S., Furukawa, Y., Snavely, N., Simon, I., Curless, B., Seitz, S.M.,
  Szeliski, R.: Building rome in a day. Communications of the ACM
  \textbf{54}(10),  105--112 (2011)

\bibitem{chang2018psmnet}
Chang, J.R., Chen, Y.S.: Pyramid stereo matching network. In: Proceedings of
  the IEEE Conference on Computer Vision and Pattern Recognition. pp.
  5410--5418 (2018)

\bibitem{Cordts2016Cityscapes}
Cordts, M., Omran, M., Ramos, S., Rehfeld, T., Enzweiler, M., Benenson, R.,
  Franke, U., Roth, S., Schiele, B.: The cityscapes dataset for semantic urban
  scene understanding. In: Proc. of the IEEE Conference on Computer Vision and
  Pattern Recognition (CVPR) (2016)

\bibitem{flownet2015dosovitskiy}
Dosovitskiy, A., Fischer, P., Ilg, E., Hausser, P., Hazirbas, C., Golkov, V.,
  van~der Smagt, P., Cremers, D., Brox, T.: Flownet: Learning optical flow with
  convolutional networks. In: Proceedings of the IEEE International Conference
  on Computer Vision. pp. 2758--2766 (2015)

\bibitem{depthnormal2015eigen}
Eigen, D., Fergus, R.: Predicting depth, surface normals and semantic labels
  with a common multi-scale convolutional architecture. In: Proceedings of the
  IEEE International Conference on Computer Vision. pp. 2650--2658 (2015)

\bibitem{depth2014eigen}
Eigen, D., Puhrsch, C., Fergus, R.: Depth map prediction from a single image
  using a multi-scale deep network. In: Advances in neural information
  processing systems. pp. 2366--2374 (2014)

\bibitem{lsdslam2014engel}
Engel, J., Sch{\"o}ps, T., Cremers, D.: Lsd-slam: Large-scale direct monocular
  slam. In: European Conference on Computer Vision. pp. 834--849. Springer
  (2014)

\bibitem{virtualkittiCVPR2016}
Gaidon, A., Wang, Q., Cabon, Y., Vig, E.: Virtual worlds as proxy for
  multi-object tracking analysis. In: CVPR (2016)

\bibitem{unsupervised2016garg}
Garg, R., BG, V.K., Carneiro, G., Reid, I.: Unsupervised cnn for single view
  depth estimation: Geometry to the rescue. In: European Conference on Computer
  Vision. pp. 740--756. Springer (2016)

\bibitem{kitti2013GeigerIJRR}
Geiger, A., Lenz, P., Stiller, C., Urtasun, R.: Vision meets robotics: The
  kitti dataset. International Journal of Robotics Research (IJRR)  (2013)

\bibitem{leftright2017godard}
Godard, C., Mac~Aodha, O., Brostow, G.J.: Unsupervised monocular depth
  estimation with left-right consistency. In: CVPR. vol.~2, p.~7 (2017)

\bibitem{displets2015guney}
Guney, F., Geiger, A.: Displets: Resolving stereo ambiguities using object
  knowledge. In: Proceedings of the IEEE Conference on Computer Vision and
  Pattern Recognition. pp. 4165--4175 (2015)

\bibitem{he2016deep}
He, K., Zhang, X., Ren, S., Sun, J.: Deep residual learning for image
  recognition. In: Proceedings of the IEEE conference on computer vision and
  pattern recognition. pp. 770--778 (2016)

\bibitem{sgmstereo2008hirschmuller}
Hirschmuller, H.: Stereo processing by semiglobal matching and mutual
  information. IEEE Transactions on pattern analysis and machine intelligence
  \textbf{30}(2),  328--341 (2008)

\bibitem{hosni2009local}
Hosni, A., Bleyer, M., Gelautz, M., Rhemann, C.: Local stereo matching using
  geodesic support weights. In: Image Processing (ICIP), 2009 16th IEEE
  International Conference on. pp. 2093--2096. IEEE (2009)

\bibitem{hosni2013fast}
Hosni, A., Rhemann, C., Bleyer, M., Rother, C., Gelautz, M.: Fast cost-volume
  filtering for visual correspondence and beyond. IEEE Transactions on Pattern
  Analysis and Machine Intelligence  \textbf{35}(2),  504--511 (2013)

\bibitem{kanade1994stereo}
Kanade, T., Okutomi, M.: A stereo matching algorithm with an adaptive window:
  Theory and experiment. IEEE transactions on pattern analysis and machine
  intelligence  \textbf{16}(9),  920--932 (1994)

\bibitem{endtoend2017kendall}
Kendall, A., Martirosyan, H., Dasgupta, S., Henry, P., Kennedy, R., Bachrach,
  A., Bry, A.: End-to-end learning of geometry and context for deep stereo
  regression. CoRR, vol. abs/1703.04309  (2017)

\bibitem{kingma2014adam}
Kingma, D.P., Ba, J.: Adam: A method for stochastic optimization. arXiv
  preprint arXiv:1412.6980  (2014)

\bibitem{kolmogorov2001graphcut}
Kolmogorov, V., Zabih, R.: Computing visual correspondence with occlusions
  using graph cuts. In: Computer Vision, 2001. ICCV 2001. Proceedings. Eighth
  IEEE International Conference on. vol.~2, pp. 508--515. IEEE (2001)

\bibitem{semidepth2017kuznietsov}
Kuznietsov, Y., St{\"u}ckler, J., Leibe, B.: Semi-supervised deep learning for
  monocular depth map prediction. In: Proc. of the IEEE Conference on Computer
  Vision and Pattern Recognition. pp. 6647--6655 (2017)

\bibitem{depthnormalcrf2015li}
Li, B., Shen, C., Dai, Y., van~den Hengel, A., He, M.: Depth and surface normal
  estimation from monocular images using regression on deep features and
  hierarchical crfs. In: Proceedings of the IEEE Conference on Computer Vision
  and Pattern Recognition. pp. 1119--1127 (2015)

\bibitem{liang2018iresnet}
Liang, Z., Feng, Y., Chen, Y.G.H.L.W., Zhang, L.Q.L.Z.J.: Learning for
  disparity estimation through feature constancy. In: Proceedings of the IEEE
  Conference on Computer Vision and Pattern Recognition. pp. 2811--2820 (2018)

\bibitem{liu2016learning}
Liu, F., Shen, C., Lin, G., Reid, I.: Learning depth from single monocular
  images using deep convolutional neural fields. IEEE transactions on pattern
  analysis and machine intelligence  \textbf{38}(10),  2024--2039 (2016)

\bibitem{discretecontinuous2014liu}
Liu, M., Salzmann, M., He, X.: Discrete-continuous depth estimation from a
  single image. In: Computer Vision and Pattern Recognition (CVPR), 2014 IEEE
  Conference on. pp. 716--723. IEEE (2014)

\bibitem{lrelu-maas2013rectifier}
Maas, A.L., Hannun, A.Y., Ng, A.Y.: Rectifier nonlinearities improve neural
  network acoustic models. In: Proc. icml. vol.~30, p.~3 (2013)

\bibitem{dispnet2016mayer}
Mayer, N., Ilg, E., Hausser, P., Fischer, P., Cremers, D., Dosovitskiy, A.,
  Brox, T.: A large dataset to train convolutional networks for disparity,
  optical flow, and scene flow estimation. In: Proceedings of the IEEE
  Conference on Computer Vision and Pattern Recognition. pp. 4040--4048 (2016)

\bibitem{orbslam2015mur}
Mur-Artal, R., Montiel, J.M.M., Tardos, J.D.: Orb-slam: a versatile and
  accurate monocular slam system. IEEE Transactions on Robotics
  \textbf{31}(5),  1147--1163 (2015)

\bibitem{dtam2011newcombe}
Newcombe, R.A., Lovegrove, S.J., Davison, A.J.: Dtam: Dense tracking and
  mapping in real-time. In: Computer Vision (ICCV), 2011 IEEE International
  Conference on. pp. 2320--2327. IEEE (2011)

\bibitem{cascade2017pang}
Pang, J., Sun, W., Ren, J., Yang, C., Yan, Q.: Cascade residual learning: A
  two-stage convolutional neural network for stereo matching. In: International
  Conf. on Computer Vision-Workshop on Geometry Meets Deep Learning (ICCVW
  2017). vol.~3 (2017)

\bibitem{ILSVRC15}
Russakovsky, O., Deng, J., Su, H., Krause, J., Satheesh, S., Ma, S., Huang, Z.,
  Karpathy, A., Khosla, A., Bernstein, M., Berg, A.C., Fei-Fei, L.: {ImageNet
  Large Scale Visual Recognition Challenge}. International Journal of Computer
  Vision (IJCV)  \textbf{115}(3),  211--252 (2015).
  \doi{10.1007/s11263-015-0816-y}

\bibitem{saxena2006learning}
Saxena, A., Chung, S.H., Ng, A.Y.: Learning depth from single monocular images.
  In: Advances in neural information processing systems. pp. 1161--1168 (2006)

\bibitem{saxena20083}
Saxena, A., Chung, S.H., Ng, A.Y.: 3-d depth reconstruction from a single still
  image. International journal of computer vision  \textbf{76}(1),  53--69
  (2008)

\bibitem{saxena2009make3d}
Saxena, A., Sun, M., Ng, A.Y.: Make3d: Learning 3d scene structure from a
  single still image. IEEE transactions on pattern analysis and machine
  intelligence  \textbf{31}(5),  824--840 (2009)

\bibitem{scharstein2002taxonomy}
Scharstein, D., Szeliski, R.: A taxonomy and evaluation of dense two-frame
  stereo correspondence algorithms. International journal of computer vision
  \textbf{47}(1-3),  7--42 (2002)

\bibitem{vgg16Simonyan14c}
Simonyan, K., Zisserman, A.: Very deep convolutional networks for large-scale
  image recognition. CoRR  \textbf{abs/1409.1556} (2014)

\bibitem{sun2003stereobeliefpropagtion}
Sun, J., Zheng, N.N., Shum, H.Y.: Stereo matching using belief propagation.
  IEEE Transactions on pattern analysis and machine intelligence
  \textbf{25}(7),  787--800 (2003)

\bibitem{tao2013depth}
Tao, M.W., Hadap, S., Malik, J., Ramamoorthi, R.: Depth from combining defocus
  and correspondence using light-field cameras. In: Computer Vision (ICCV),
  2013 IEEE International Conference on. pp. 673--680. IEEE (2013)

\bibitem{unsupervised2017tonioni}
Tonioni, A., Poggi, M., Mattoccia, S., Di~Stefano, L.: Unsupervised adaptation
  for deep stereo. In: Proceedings of the IEEE Conference on Computer Vision
  and Pattern Recognition. pp. 1605--1613 (2017)

\bibitem{torr1999feature}
Torr, P.H., Zisserman, A.: Feature based methods for structure and motion
  estimation. In: International workshop on vision algorithms. pp. 278--294.
  Springer (1999)

\bibitem{demon2017ummenhofer}
Ummenhofer, B., Zhou, H., Uhrig, J., Mayer, N., Ilg, E., Dosovitskiy, A., Brox,
  T.: Demon: Depth and motion network for learning monocular stereo. In: IEEE
  Conference on Computer Vision and Pattern Recognition (CVPR). vol.~5 (2017)

\bibitem{sfmnetarxiv2017vijayanarasimhan}
Vijayanarasimhan, S., Ricco, S., Schmid, C., Sukthankar, R., Fragkiadaki, K.:
  Sfm-net: Learning of structure and motion from video. arXiv preprint
  arXiv:1704.07804  (2017)

\bibitem{learningdirectarxiv2017wang}
Wang, C., Buenaposada, J.M., Zhu, R., Lucey, S.: Learning depth from monocular
  videos using direct methods. arXiv preprint arXiv:1712.00175  (2017)

\bibitem{wu2011visualsfm}
Wu, C., et~al.: Visualsfm: A visual structure from motion system  (2011)

\bibitem{multi2017danxu}
Xu, D., Ricci, E., Ouyang, W., Wang, X., Sebe, N.: Multi-scale continuous crfs
  as sequential deep networks for monocular depth estimation. In: Proceedings
  of CVPR (2017)

\bibitem{yang2010constantspacebp}
Yang, Q., Wang, L., Ahuja, N.: A constant-space belief propagation algorithm
  for stereo matching. In: Computer vision and pattern recognition (CVPR), 2010
  IEEE Conference on. pp. 1458--1465. IEEE (2010)

\bibitem{yang2017unsupervisedarxiv}
Yang, Z., Wang, P., Xu, W., Zhao, L., Nevatia, R.: Unsupervised learning of
  geometry with edge-aware depth-normal consistency. arXiv preprint
  arXiv:1711.03665  (2017)

\bibitem{mccnn2015zbontar}
Zbontar, J., LeCun, Y.: Computing the stereo matching cost with a convolutional
  neural network. In: Proceedings of the IEEE conference on computer vision and
  pattern recognition. pp. 1592--1599 (2015)

\bibitem{unsupervised2017zhou}
Zhou, C., Zhang, H., Shen, X., Jia, J.: Unsupervised learning of stereo
  matching. In: Proceedings of the IEEE Conference on Computer Vision and
  Pattern Recognition. pp. 1567--1575 (2017)

\bibitem{sfmlearner2017zhou}
Zhou, T., Brown, M., Snavely, N., Lowe, D.G.: Unsupervised learning of depth
  and ego-motion from video. In: CVPR. vol.~2, p.~7 (2017)

\end{thebibliography}

\clearpage
\includepdf[pages=1]{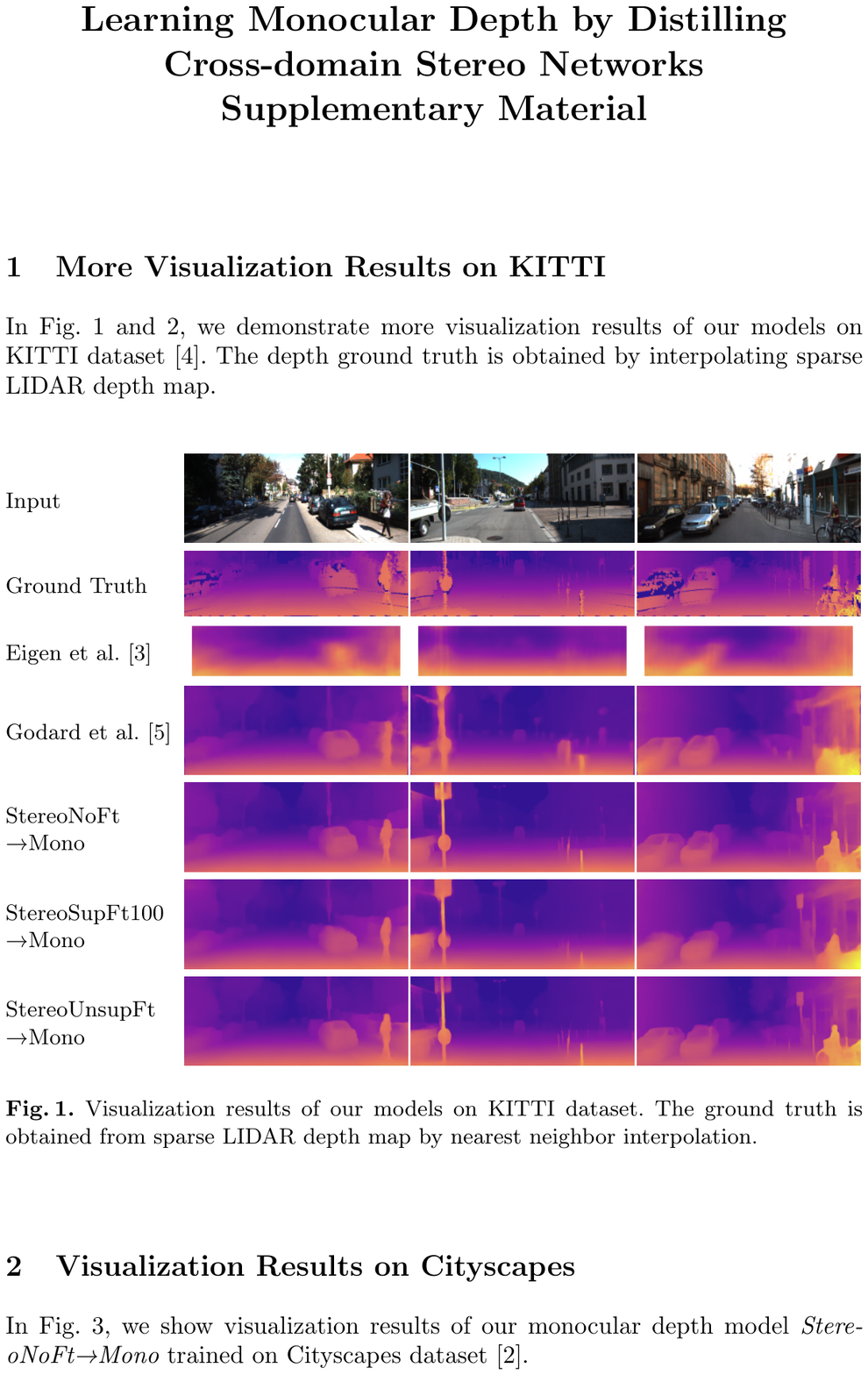}
\clearpage
\includepdf[pages=2]{supp.pdf}
\clearpage
\includepdf[pages=3]{supp.pdf}
\clearpage
\includepdf[pages=4]{supp.pdf}

\end{document}